\documentclass[twocolumn,11pt]{IEEEtran}
\begin{document}
%
\title{Diagnosability of Fuzzy Discrete Event Systems\thanks{This work was supported in part by the National
Natural Science Foundation under Grant 90303024 and Grant
60573006, the Higher School Doctoral Subject Foundation of
Ministry of Education under Grant 20050558015, and the Guangdong
Province Natural Science Foundation under Grant 020146 and Grant
031541 of China. }}
\author{Fuchun Liu$^{a,b}$, Daowen Qiu$^{a}$
\thanks{Corresponding author (D. Qiu).}\thanks{E-mail addresses:
issqdw@mail.sysu.edu.cn (D. Qiu); liufch@gdut.edu.cn (F. Liu)}, Hongyan Xing$^{a,b}$, and Zhujun Fan$^{a}$\\
{\footnotesize   $^{a}$Department of Computer Science, Zhongshan
University,
Guangzhou 510275, China}\\
{\footnotesize  $^{b}$Faculty of Applied Mathematics, Guangdong
University of Technology, Guangzhou 510090, China} }

\date{}
\markboth{Diagnosability of Fuzzy Discrete Event Systems}{Shell
\MakeLowercase{\textit{et al.}}: Bare Demo of IEEEtran.cls for
Journals}
\maketitle

\begin{abstract}
In order to more effectively cope with the real-world problems of
vagueness, {\it fuzzy discrete event systems} (FDESs) were
proposed recently, and the supervisory control theory of FDESs was
developed. In view of the importance of failure diagnosis, in this
paper, we present an approach of the failure diagnosis in the
framework of FDESs. More specifically: (1) We formalize the
definition of diagnosability for FDESs, in which the observable
set and failure set of events are {\it fuzzy}, that is, each event
has certain degree to be observable and unobservable, and, also,
each event may possess different possibility of failure occurring.
(2) Through the construction of observability-based diagnosers of
FDESs, we investigate its some basic properties. In particular, we
present a necessary and sufficient condition for diagnosability of
FDESs.  (3) Some examples serving to illuminate the applications
of the diagnosability of FDESs are described. To conclude, some
related issues are raised for further consideration.
\end{abstract}

\IEEEpeerreviewmaketitle

\section{Introduction}
A {\it discrete event system} (DES) is a dynamical system whose
state space is discrete and whose states can only change as a
result of asynchronous occurrence of instantaneous events over
time.  Up to now, DESs have been successfully applied to many
engineering fields [4].  In most of engineering applications, the
states of a DES are crisp. However, this is not the case in many
other applications in complex systems such as biomedical systems
and economic systems. For example, it is vague when a man's
condition of the body is said to be ``good". Moreover, it is
imprecise to say at what point exactly a man has changed from
state ``good" to state ``poor".  Therefore, Lin and Ying [18,19]
initiated significantly the study of {\it fuzzy discrete event
systems} (FDESs) by combining fuzzy set theory with crisp DESs.
Notably, FDESs have been applied to biomedical control for
HIV/AIDS treatment planning [20,21]. And R. Huq {\it et al} have
presented a novel intelligent sensory information processing using
FDESs for robotic control recently [10, 11].

As Lin and Ying [19] pointed out, a comprehensive theory of FDESs
still needs to be set up, including many important concepts,
methods and theorems, such as controllability, observability, and
optimal control. These issues have been partially investigated in
[2, 3, 28]. Qiu [28] established the supervisory control theory of
FDESs, and found a method of checking the existence of supervisors
for FDESs; and independently, Cao and Ying [2, 3] significantly
developed FDESs. As a continuation, this paper is to deal with the
failure diagnosis for FDESs.

It is well known that the issues of diagnosability for DESs are of
practical and theoretical importance, and have received extensive
attention in recent years [5-9,12,13,15-17,23-27,29-39]. However,
the observability and the failure set of events in the literature
are usually {\it crisp}. Motivated by the fuzziness of
observability for some events in real-life situation, in this
paper,  the observable set and failure set of events are {\it
fuzzy}. That is, each event has certain degree to be observable
and unobservable, and, also, each event may possess different
possibility of failure occurring. We formalize the definition of
diagnosability for FDESs using the fuzzy observable set and the
fuzzy failure set of events.

Generally speaking, a fuzzy language generated by a fuzzy finite
automaton is said to be diagnosable if, based on the degree of
observability and the possibility of failure occurring on events,
the occurrence of failures can be always detected within a finite
delay according to the observed information of the traces. Through
the construction of observability-based diagnosers of FDESs, we
investigate some basic properties concerning the diagnosers. In
particular, we present a necessary and sufficient condition for
diagnosability of FDESs, that is, a fuzzy language is
$F_{i}$-diagnosable if and only if there are no
$F_{i}$-indeterminate cycles in the diagnoser with respect to each
event. Our results may better deal with the problems of fuzziness,
impreciseness and subjectivity in the failure diagnosis, and,
generalize the important consequences in classical DESs introduced
by Sampath {\it et al} in their seminal works [31, 32]. In order
to illustrate the applications of the diagnosability of FDESs,
some examples are provided to illuminate the results derived.

 This paper is organized as follows. Section II recalls some preliminaries
and notations concerning FDESs. In Section III, an approach to
defining diagnosability for FDESs is presented. In Section IV, we
construct the observability-based diagnosers of FDESs, and some
main properties of the diagnosers are investigated. In particular,
we present a necessary and sufficient condition for diagnosability
of FDESs. Finally, some examples are provided to illustrate the
condition of diagnosability for FDESs in Section V. To conclude,
in Section VI, we summarize the main results of the paper and
address some related issues.

\section{Preliminaries}\label{S:Prepare}
In this section, we briefly recall some preliminaries regarding
fuzzy finite automata. For a detailed introduction, we may refer
to [18, 19, 28].

In the setting of FDESs, a fuzzy state is represented as a vector
$[a_{1}, a_{2},\cdots, a_{n}]$, which stands for the possibility
distributions over crisp states, that is, $a_{i}\in [0,1]$
represents the possibility that the system is in the $i$th crisp
state, ($i=1,2,\cdots,n$). Similarly, a fuzzy event is denoted by
a matrices $\sigma=[a_{ij}]_{n\times n}$, and $a_{ij}\in [0,1]$
means the possibility for the system to transfer from the $i$th
crisp state to the $j$th crisp state when event $\sigma$ occurs,
and $n$ is the number of all possible crisp states. Hence, a fuzzy
finite automaton is defined as follows.

{\it Definition 1 [28]:}  A {\it  fuzzy finite automaton} is a
fuzzy system $$G=(Q,\Sigma,\delta, q_{0}),$$ where $Q$ is the set
of some state vectors (fuzzy states) over crisp state set; $q_{0}$
is the initial fuzzy state; $\Sigma$ is the set of matrices (fuzzy
events); $\delta: Q\times\Sigma\rightarrow Q $ is a transition
function which is defined by $\delta(q,\sigma)= q\odot \sigma$ for
$q\in Q$ and $\sigma\in\Sigma$, where $\odot$ denotes the {\it
max-min} operation in fuzzy set theory [14].

{\it Remark 1:} The transition function $\delta$ can be naturally
extended to $Q\times\Sigma^{*}$ in the following manner:
$$\delta(q,\epsilon)= q,\hskip 3mm\delta(q,s\sigma)=\delta(\delta(q,s),\sigma),$$ where
$\Sigma^{*}$ is the Kleene closure of $\Sigma$, $\epsilon$ denotes
the empty string, $q\in Q$, $\sigma\in\Sigma$ and
$s\in\Sigma^{*}$. Moreover, $\delta$ can be regarded as a partial
transition function in practice. In biomedical engineering [20],
for example, although many treatments (fuzzy events) are available
for a patient, but in fact, only one or a few treatments are
adopted by doctors according to the patient's conditions (fuzzy
states). We can see Example 2 later for details.

The fuzzy languages generated by $G$ is denoted by ${\cal L}_{G}$
or ${\cal L}$ for simplicity [28], which is a function from
$\Sigma^{*}$ to $[0,1]$. Let $s\in \Sigma^{*}$. The postlanguage
of ${\cal L}$ after $s$ is the set of continuations of $s$ in all
physically possible traces, i.e., $$ {\cal L}/s=\{t \in
\Sigma^{*}: (\exists q\in Q )[\delta(q_{0}, st)= q \wedge {\cal
L}(st)>0]\}.
$$

From [18, 19, 28], we know that each fuzzy event is associated
with a degree of controllability, so, the uncontrollable set
$\widetilde{\Sigma}_{uc}$ and controllable set
$\widetilde{\Sigma}_{c}$ are two fuzzy subsets of $\Sigma$, and
satisfy: for any $ \sigma \in \widetilde{\Sigma}$,
$$
\widetilde{\Sigma}_{uc}(\sigma)+\widetilde{\Sigma}_{c}(\sigma)=1.
$$
Analogously, we think that each fuzzy event is associated with a
degree of observability. For instance, for some treatments (fuzzy
events) in biomedical systems modelled by a fuzzy finite
automaton, some effects are observable (headache disappears, for
example), but some are unobservable (for instance, some potential
side effects of treatment). Therefore, the unobservable set
$\widetilde{\Sigma}_{uo}$ and observable set
$\widetilde{\Sigma}_{o}$ are two fuzzy subsets of
$\widetilde{\Sigma}$, too, and satisfy: for any $ \sigma \in
\widetilde{\Sigma}$,
\begin{equation}
\widetilde{\Sigma}_{uo}(\sigma)+\widetilde{\Sigma}_{o}(\sigma)=1.
\end{equation}
Furthermore, we define $\widetilde{\Sigma}_{o}(\epsilon)=0$, and
\begin{equation}
\widetilde{\Sigma}_{o}(s)=\min\{\widetilde{\Sigma}_{o}
(\sigma_{i}):i=1,2,\ldots,m\}\end{equation} for
$s=\sigma_{1}\sigma_{2}\ldots\sigma_{m}\in \Sigma^{*}$.

We define the maximal observable set $\Sigma_{mo}$, which is
composed of the events that have the greatest degree of
observability among $\Sigma$, i.e.,
\begin{equation}
\Sigma_{mo}= \{\sigma\in \Sigma: (\forall a\in
\Sigma)[\widetilde{\Sigma}_{o}(\sigma)\geq
\widetilde{\Sigma}_{o}(a)]\}.
\end{equation}
Let ${\cal L}_{G}(q)$ is the set of all traces that originate from
fuzzy state $ q $. Denote
\begin{equation}
\begin{array}{ll}
{\cal L}_{1}(q, \sigma )&=\{a \in \Sigma\cap{\cal L}_{G}(q):\\
&(a\in\Sigma_{mo})\vee[\widetilde{\Sigma}_{o}(a)>\widetilde{\Sigma}_{o}(\sigma)
]\},
\end{array}
\end{equation}
\begin{equation}
\begin{array}{ll}
{\cal L}_{2}(q, \sigma)&= \{ua\in {\cal L}_{G}(q):(\parallel u
\parallel\geq 1)\\ &\wedge[\widetilde{\Sigma}_{o}(\sigma)\geq
\widetilde{M}_{o}(u)]\wedge [a\in {\cal L}_{1}(q, \sigma )]\},
\end{array}
\end{equation}
where $\parallel u\parallel$ denotes the length of string $u$, and
$\widetilde{M}_{o}(u)= \max\{\widetilde{\Sigma}_{o}(\sigma):
\sigma\in u\}$. Intuitively, ${\cal L}_{1}(q, \sigma )$ collects
all of single fuzzy event whose degree of observability is
 either the greatest among $\Sigma$ or greater than
 $\widetilde{\Sigma}_{o}(\sigma)$.
 And ${\cal L}_{2}(q, \sigma )$ consists of the strings $ua$
 containing at least two fuzzy events, in which
 the degree of observability for any event of $u$ is less than
 or equal to that of $\sigma$ and $a\in {\cal L}_{1}(q, \sigma )$. We denote
\begin{equation}
{\cal L}(q, \sigma)= {\cal L}_{1}(q, \sigma )\cup {\cal L}_{2}(q,
\sigma ),
\end{equation}
\begin{equation}
{\cal L}_{a}(q, \sigma )= \left\{s \in {\cal L}(q, \sigma): s_{f}
= a\right\},
\end{equation}
where ${\cal L}_{a}(q, \sigma )$ represents those strings in
${\cal L}(q, \sigma)$ that end with event $a$.

\section{Approaches to Defining Diagnosability for FDESs}\label{S:OFD}
In this section, we will give a definition of the diagnosability
for FDESs using the fuzzy observable set $\widetilde{\Sigma}_{o}$
and the fuzzy failure set $\widetilde{\Sigma}_{f}$.

As mentioned above, in biomedical systems modelled by a fuzzy
finite automaton, some effects are observable, but some are
unobservable, even some effects are undesired failures (for
example, some potential side effects). Therefore, in the setting
of FDESs, the failure set of events, as a subset of the
unobservable set $\widetilde{\Sigma}_{uo}$, is also regarded as a
fuzzy subset of $\Sigma$. We denote it as
$\widetilde{\Sigma}_{f}$, and, for each fuzzy event
$\sigma\in\Sigma$, $\widetilde{\Sigma}_{f}(\sigma)$ represents the
possibility of the failure occurring on $\sigma$. Since diagnosis
is generally based on the unobservable failures [31,32,36],
without loss of generality, we can assume that
$\widetilde{\Sigma}_{f}\widetilde{\subseteq}
\widetilde{\Sigma}_{uo}$, that is,
$\widetilde{\Sigma}_{f}(\sigma)\leq
\widetilde{\Sigma}_{uo}(\sigma)$ for any $\sigma\in\Sigma$, which
means that failures are always unobservable.

Usually, the failure set $\widetilde{\Sigma}_{f}$ is partitioned
into a set of failure types $f_{1}, f_{2},\ldots,f_{m}$, i.e.,
\begin{equation}
\widetilde{\Sigma}_{f}=\widetilde{\Sigma}_{f_{1}}
\widetilde{\cup}\widetilde{\Sigma}_{f_{2}}\widetilde{\cup}
\ldots\widetilde{\cup}\widetilde{\Sigma}_{f_{m}}\end{equation}
 where
$\widetilde{\cup}$ is Zadeh fuzzy OR operator [14], that is,
$$\widetilde{\Sigma}_{f}(\sigma)=
\max\left\{\widetilde{\Sigma}_{f_{i}}(\sigma):
i=1,2,\ldots,m\right\}$$ for any $ \sigma\in \Sigma^{*}$. Let
$s_{f}$ denote the final fuzzy event of $s\in \Sigma^{*}$. We
define
\begin{equation}
\begin{array}{ll}
\Psi_{\sigma}(\widetilde{\Sigma}_{f_{i}})&=\{s \in
\Sigma^{*}: (\exists q\in Q )[\delta(q_{0}, s)= q]\\
 &\wedge [{\cal
L}(s)>0] \wedge [\widetilde{\Sigma}_{f_{i}}(s_{f})\geq
\widetilde{\Sigma}_{f_{i}}(\sigma)]\}. \end{array}
\end{equation}
Intuitively, $ \Psi_{\sigma}(\widetilde{\Sigma}_{f_{i}})$ is the
set of all physically possible traces that end in a event on which
the possibility of failure of type $f_{i}$ occurring is not less
than $\widetilde{\Sigma}_{f_{i}}(\sigma)$.

When a string of events occurs in a system, the events sequence is
filtered by a projection based on their degrees of observability.

{\it Definition 2:}  For $\sigma\in \Sigma $, the {\it
$\sigma$-projection} $P_{\sigma}:\Sigma^{*}\rightarrow \Sigma^{*}$
is defined as: For any $ a\in \Sigma $ and $ s\in \Sigma^{*}$,
\begin{equation}
P_{\sigma}(a)=\left\{
\begin{array}{ll}
a,    &{\rm if}  \hskip 2mm a\in \Sigma_{mo} \hskip 2mm{\rm or}
\hskip 1mm\widetilde{\Sigma}_{o}(a)>\widetilde{\Sigma}_{o}(\sigma)
,\\
\epsilon, &{\rm otherwise},
\end{array}
\right.
\end{equation}
and $ P_{\sigma}(\epsilon)=\epsilon$, \hskip 2mm
$P_{\sigma}(sa)=P_{\sigma}(s)P_{\sigma}(a)$.

The inverse projection operator is given by:
$$\begin{array}{ll}
P_{\sigma}^{-1}(y)&=\{s \in \Sigma^{*}: (\exists q\in Q
)\\&[\delta(q_{0}, s)= q]\wedge[{\cal
L}(s)>0]\wedge[P_{\sigma}(s)=y] \}.\end{array}
$$

The purpose of $\sigma $-projection is to erase the events whose
degree of observability is not greater than
$\widetilde{\Sigma}_{o}(\sigma)$ in a string. Especially, when a
deterministic or nondeterministic finite automaton is regarded as
a special form of fuzzy finite automaton, then all $\sigma
$-projections are equal, and, all of them degenerate to projection
$P:\Sigma^{*}\rightarrow \Sigma_{o}^{*}$ in the usual manner,
which simply erases the unobservable events [31, 32].

{\it Remark 2:} In order to avoid the case that the event set of
the diagnoser constructed later is null, we introduce the maximal
observable set $\Sigma_{mo}$ in the definition of
$\sigma$-projection $P_{\sigma}$, since it is impossible to
diagnose the failure using a diagnoser with a null event set.

For the sake of simplicity, we make the following two assumptions
about the fuzzy automaton $G$, which are similar to those in [31,
32, 36].

({\it A1}): Language ${\cal L}_{G}$ is live. This means that
system cannot reach a state without transitions.

({\it A2}): For any $\sigma\in \Sigma$ and state $q\in Q$, there
exists $n_{0}\in N $ such that $\parallel t
\parallel\leq n_{0}$ for every $t\in {\cal L}(q, \sigma)$.

Intuitively, assumption ({\it A1}) indicates that there is a
transition defined at each state, and ({\it A2}) means that for
any event $\sigma\in \Sigma$, before generating an event whose
observability degree is the greatest among $\Sigma$ or greater
than $\widetilde{\Sigma}_{o}(\sigma)$,  $G$ does not generate
arbitrarily long sequences in which each event's degree of
observability is less than $\widetilde{\Sigma}_{o}(\sigma)$.

In order to compare diagnosability for FDESs with that for
classical DESs, we recall the definition of diagnosability for
classical DESs presented by Sampath {\it et al} [31].

{\it Definition 3 [31]:} A language $L$ are said to be {\it
diagnosable} with respect to the projection $P$ and the partition
$\Pi_{f}$ on $\Sigma_{f}$, if the following holds:
\begin{equation}\begin{array}{l} (\forall i\in\Pi_{f})(\exists
n_{i}\in {\bf N})[\forall s\in \Psi(\Sigma_{f_{i}})]\\(\forall
t\in L/s)[\parallel t
\parallel\geq
n_{i}\Rightarrow D)
\end{array}
\end{equation}
where the diagnosability condition function $D$ is
\begin{equation}
\omega\in P^{-1}[P(st)] \Rightarrow\Sigma_{f_{i}}\in \omega.
\end{equation}

The objective of diagnosis for classical DESs is to detect the
unobservable failures from the record of the observed events. As
mentioned above, in FDESs, the failures may occur on every fuzzy
event, only their possibilities of failure occurring are
different. Therefore, the purpose of diagnosis for FDESs is to
detect the failures from the sequence of the observed events,
based on the degree of observability and the possibility of
failure occurring. Now let us give the definition of
diagnosability for FDESs.

{\it Definition 4:} Let ${\cal L}$ be a language generated by a
fuzzy finite automaton $G=(Q,\Sigma,\delta, q_{0})$ and $\sigma\in
\Sigma$. ${\cal L}$ is said to be {\it$ F_{i}$-diagnosable with
respect to $\sigma$}, if there exists $n_{i}\in N $ such that for
any $s\in \Psi_{\sigma}(\widetilde{\Sigma}_{f_{i}})$ and any $t\in
{\cal L}/s$ where $\parallel t \parallel\geq n_{i}$, the following
holds:
\begin{equation}
\widetilde{\Sigma}_{f_{i}}(\sigma)\leq
\min\left\{\widetilde{\Sigma}_{f_{i}}(\omega): \omega \in
P_{\sigma}^{-1}(P_{\sigma}(st))\right\}.
\end{equation}

Denote $ \Sigma_{fail_{i}}= \left\{\sigma\in \Sigma:
\widetilde{\Sigma}_{f_{i}}(\sigma)>0\right\}$.  If for each
$\sigma\in\Sigma_{fail_{i}}$, ${\cal L}$ is $ F_{i}$-diagnosable
with respect to $\sigma$, then ${\cal L}$ is said to be {\it $
F_{i}$-diagnosable}.

Intuitively, ${\cal L}$ being $ F_{i}$-diagnosable with respect to
$\sigma$ means that, for any physically possible trace $s$ where
the possibility that failure of type $f_{i}$ occurs on $s_{f}$ is
not less than that on $\sigma$, any sufficiently long continuation
$t$ of $s$, and any trace $\omega$, if $\omega$ produces the same
record by the $\sigma $-projection as the trace $st$, then the
possibility that failure of type $f_{i}$ occurs on $\omega$ must
be not less than that on $\sigma$, too. In other words, if the
failure type $f_{i}$ has occurred on event $s_{f}$, then $f_{i}$
must also occur on every trace $\omega$ whose observed record is
the same as $st$.

{\it Remark 3:} If the observability and possibility of failure
occurring of each event are crisp, i.e.,
$\widetilde{\Sigma}_{o}(\sigma)$,
$\widetilde{\Sigma}_{f_{i}}(\sigma)\in \{0,1\}$, then the
definition of diagnosability for FDESs reduces to Definition 3,
the diagnosability for classical DESs presented by Sampath {\it et
al} [31].

We present an example to explain the definition of diagnosability
for FDESs, and the real-world application example will be given in
Example 2 later.

{\it Example 1.} Consider the fuzzy automaton $G=(Q, \Sigma,
\delta, q_{0})$ represented in Fig.1,

\setlength{\unitlength}{0.1cm}
\begin{picture}(90,38)

\put(5,27){\circle{5}\makebox(-10,0){$q_{0}$}}
\put(20,27){\circle{5}\makebox(-10,0){$q_{1}$}}
\put(35,27){\circle{5}\makebox(-10,0){$q_{2}$}}
\put(50,27){\circle{5}\makebox(-10,0){$q_{3}$}}
\put(35,11){\circle{5}\makebox(-10,0){$q_{4}$}}

\put(8,27){\vector(1,0){9}} \put(23,27){\vector(1,0){9}}
\put(38,27){\vector(1,0){9}} \put(35,24.3){\vector(0,-1){10}}

\qbezier(37,29)(45,39)(50,30) \put(48.7,31.3){\vector(1,-1){2}}

\put(56.7,27){\circle{8}} \put(61,27){\vector(0,1){1}}
\put(41.8,11){\circle{8}} \put(46,11){\vector(0,1){1}}

\put(11,29){\makebox(0,0)[c]{$\alpha$}}
\put(28,29){\makebox(0,0)[c]{$\beta$}}
\put(42,29){\makebox(0,0)[c]{$\beta$}}
\put(64,27){\makebox(0,0)[c]{$\theta$}}
\put(37,19){\makebox(0,0)[c]{$\gamma$}}
\put(48,11){\makebox(0,0)[c]{$\theta$}}
\put(49,35){\makebox(0,0)[c]{$\tau$}}

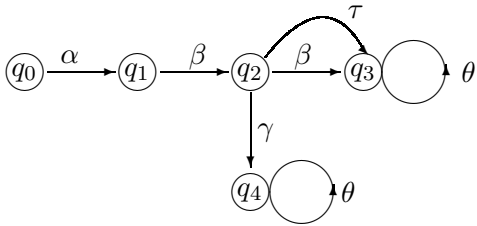
\put(23,1){\makebox(20,1)[c]{{\footnotesize Fig.1. The fuzzy
automaton of Example 1. }}}
\end{picture}
\\ where $Q=\{q_{0}, q_{1},\ldots, q_{4}\}$, $q_{0}=[0.8,0.2]$,
and $\Sigma=\{\alpha, \beta, \gamma, \tau, \theta\}$ is defined as
follows:
$$
\begin{array}{cc}
\alpha=\left[
\begin{array}{cc}
0.8 & 0.4\\
0.4& 0.8
\end{array}
\right],
 &\beta=\left[
\begin{array}{cc}
0.4 & 0.8\\
0.8& 0.6
\end{array}
\right],\\ \gamma=\left[
\begin{array}{cc}
0.4 & 0.4\\
0.4& 0.4
\end{array}
\right],&\tau=\left[
\begin{array}{cc}
0.6 & 0.4\\
0.8& 0.6
\end{array}
\right],\\
 \theta=\left[
\begin{array}{cc}
0.9 & 0.2\\
0.2& 0.9
\end{array}
\right]. \end{array}
$$

Note that $\delta$ is defined with {\it max-min} operation, we can
calculate the other fuzzy states:
 $ q_{1}=[0.8, 0.4]$,
$q_{2}=[0.4, 0.8]$, $q_{3}=[0.8,0.6]$, and $q_{4}=[0.4,0.4]$.

Suppose that the degree of observability and the possibility of
failure occurring on each fuzzy event are defined as follows:
$$\widetilde{\Sigma}_{o}(\alpha)=0.8, \hskip 2mm \widetilde{\Sigma}_{o}(\beta)=0.5, \hskip
2mm \widetilde{\Sigma}_{o}(\gamma)=0.3,$$
$$\widetilde{\Sigma}_{o}(\theta)=0.7, \hskip 2mm
\widetilde{\Sigma}_{o}(\tau)=0.3;\hskip
2mm\widetilde{\Sigma}_{f_{1}}(\alpha)=0.2,$$
$$\widetilde{\Sigma}_{f_{1}}(\beta)=0.4,\hskip 2mm\widetilde{\Sigma}_{f_{1}}(\gamma)=0.3, \hskip 2mm
\widetilde{\Sigma}_{f_{1}}(\theta)=0.3,$$
$$\widetilde{\Sigma}_{f_{1}}(\tau)=0.6;\hskip 2mm\widetilde{\Sigma}_{f_{2}}(\alpha)=0.1, \hskip 2mm
\widetilde{\Sigma}_{f_{2}}(\beta)=0.3,$$
$$\widetilde{\Sigma}_{f_{2}}(\gamma)=0.4, \hskip 2mm
\widetilde{\Sigma}_{f_{2}}(\theta)=0.2, \hskip 2mm
\widetilde{\Sigma}_{f_{2}}(\tau)=0.5.$$ In the following, we will
use Definition 4 to  verify two conclusions: (1) the language
${\cal L}$ generated by $G$ is not $ F_{1}$-diagnosable with
respect to $\tau$, but (2) ${\cal L}$ is $ F_{2}$-diagnosable with
respect to $\beta$.

In fact, when $\sigma=\tau$, for $\forall n_{i}\in N $, we take
$s=\alpha\beta\tau$,  $t=\theta^{n_{i}+1}$, and take
$\omega=\alpha\beta\gamma\theta^{n_{i}+1}$. Obviously, $\omega \in
P_{\sigma}^{-1}(P_{\sigma}(st))$, but
$\widetilde{\Sigma}_{f_{1}}(\sigma)=0.6$, while
$\widetilde{\Sigma}_{f_{1}}(\omega)=0.4$. Therefore, Ineq.(13)
does not hold, so ${\cal L}$ is not $ F_{1}$-diagnosable with
respect to $\tau$.

When $\sigma=\beta$, we take $n_{i}=2$, then for any $s\in
\Psi_{\sigma}(\widetilde{\Sigma}_{f_{2}})$, (i.e.,$s=\alpha\beta$,
$\alpha\beta\beta$, $\alpha\beta\tau$, or $\alpha\beta\gamma$),
and any $t\in {\cal L}/s$, where $\parallel t \parallel\geq
n_{i}$, we have
$$P_{\sigma}^{-1}(P_{\sigma}(st))=
\{\alpha\beta\tau\theta^{k}, \alpha\beta\beta\theta^{k},
\alpha\beta\gamma\theta^{k}: k\geq 1 \}.$$ Due to each element in
$P_{\sigma}^{-1}(P_{\sigma}(st))$ containing $\beta$, therefore,
for any $\omega \in P_{\sigma}^{-1}(P_{\sigma}(st))$, we have
$\widetilde{\Sigma}_{f_{2}}(\sigma)\leq\widetilde{\Sigma}_{f_{2}}(\omega)$,
that is, ${\cal L}$ is $ F_{2}$-diagnosable with respect to
$\beta$.

\section{Necessary and Sufficient Condition of Diagnosability for FDESs}\label{S:NAOES}
In this section, through the construction of observability-based
diagnosers of FDESs, we investigate some main properties of the
diagnosers. In particular, we present a necessary and sufficient
condition for diagnosability of FDESs. Our results not only
generalize the significant consequences in classical DESs
introduced by Sampath {\it et al} [31], but also may better deal
with the problems of vagueness in real-world situation. Example 2
in Section V verifies this view to a certain degree.

\subsection{Construction of the Diagnosers}
 We firstly present the construction of the observability-based
diagnoser, which is a finite automaton built on fuzzy finite
automaton $G$.

Denote the set of possible failure labels as $
\triangle=\left\{N\right\}\cup 2^{\triangle_{f}} $, where $N$
stands for ``normal", and $2^{\triangle_{f}}$ denotes the power
set of $\triangle_{f}=\{F_{1},\cdots,F_{m}\}$ [31]. For $\sigma\in
\Sigma$, we define a subset of $Q$ as
\begin{equation}
\begin{array}{ll}
Q_{\sigma}&=\{q_{0}\}\cup\{q\in Q:(\exists q^{'} \in Q )(\exists a
\in\Sigma)\\&[\delta(q^{'},a)= q \wedge a\in {\cal L}_{1}(q,
\sigma )]\},\end{array}
\end{equation}
i.e., $Q_{\sigma}$ is composed of the initial state $q_{0}$ and
the states reachable from one event whose degree of observability
is either the greatest among $\Sigma$ or greater than
$\widetilde{\Sigma}_{o}(\sigma)$.

{\it Definition 5:}  Let $G=(Q,\Sigma,\delta, q_{0})$ be a fuzzy
finite automaton and $\sigma\in \Sigma_{fail_{i}}$.  The {\it
diagnoser with respect to $\sigma$} is the finite automaton
\begin{equation}
G_{d}=(Q_{d},\Sigma_{d},\delta_{d}, \chi_{0}),
\end{equation}
where the initial state $\chi_{0}= \left\{(q_{0},
\left\{N\right\})\right\}$, means that the automaton $G$ is normal
to start with.  The set of events of the diagnoser is
\begin{equation}
\Sigma_{d}=\left\{a\in \Sigma:  (a\in \Sigma_{mo})\vee
[\widetilde{\Sigma}_{o}(a)>\widetilde{\Sigma}_{o}(\sigma)]\right\}.
\end{equation}
The state space $ Q_{d}\subseteq Q_{\sigma}\times\triangle$ is
composed of the states reachable from $\chi_{0}$ under
$\delta_{d}$. A state $\chi$ of $ Q_{d}$ is of the form
\begin{equation}\chi=\left\{(q_{1}, \ell_{1}), (q_{2}, \ell_{2}),\ldots, (q_{n},
\ell_{n})\right\},\end{equation} where $q_{i}\in Q_{\sigma}$ and
$\ell_{i}\in \triangle $, i.e., $\ell_{i}$ is the form
$\ell_{i}=\left\{N\right\}$, or $\ell_{i}=\left\{F_{i_{1}},
F_{i_{2}},\ldots,F_{i_{k}} \right\}$.  And $\delta_{d}$ is the
partial transition function of the diagnoser, which will be
constructed in Definition 7.

{\it Definition 6:}  The {\it label propagation function} $LP:
Q_{\sigma}\times\triangle\times\Sigma^{*}\rightarrow\triangle$ is
defined as follows: For $q\in Q_{\sigma}, \ell\in \triangle $, and
$s\in {\cal L}(q, \sigma)$,
\begin{equation}
\begin{array}{ll}
LP(q, \ell, s)\\=\left\{
\begin{array}{ll}
\left\{N\right\},    {\rm if}  \hskip 1mm \ell=\left\{N\right\}
\hskip 1mm{\rm and}
\hskip 1mm \forall i [\widetilde{\Sigma}_{f_{i}}(s)<\widetilde{\Sigma}_{f_{i}}(\sigma)],\\
\left\{F_{i}:  F_{i}\in\ell \vee
\widetilde{\Sigma}_{f_{i}}(s)\geq\widetilde{\Sigma}_{f_{i}}(\sigma)\right\},
{\rm otherwise}.
\end{array}
\right. \end{array}
\end{equation}

 The label propagation function is due to describe the changes
of label from one state of diagnoser to another. Obviously, label
$F_{i}$ is added whenever the possibility of the $ith$ type
failure occurring on the string $s$ is not less than
$\widetilde{\Sigma}_{f_{i}}(\sigma)$, and once this label is
appended, it cannot be removed in the successor states of the
diagnoser.

{\it Definition 7:}  The {\it transition function} of the
diagnoser
 $\delta_{d}: Q_{d}\times\Sigma_{d}\rightarrow Q_{d}$ is defined
as
\begin{equation}
\delta_{d}(\chi, a)=\bigcup_{(q_{i}, \ell_{i})\in \chi}
\bigcup_{s\in {\cal L}_{a}(q_{i}, \sigma )}\left\{ (\delta(q_{i},
s), LP(q_{i}, \ell_{i}, s) )\right\}.
\end{equation}

For example, $\delta_{d}(\chi_{0}, \alpha)=\{(q_{1},\{N\})
,(q_{5}, \{F_{1}\})\}$ in Fig. 4 of Example 2.

\subsection{Some Properties of the Diagnosers}
In this subsection, we present some main properties of the
diagnoser, which will be used to prove the condition of the
diagnosability for FDESs.

{\it Property 1:}  Let $G=(Q,\Sigma,\delta, q_{0})$ be a fuzzy
finite automaton, and let $G_{d}=(Q_{d},\Sigma_{d},\delta_{d},
\chi_{0}) $ be the diagnoser with respect to $\sigma$, where
$\sigma\in \Sigma_{fail_{i}}$. For $\chi_{1}, \chi_{2}\in Q_{d}$,
$s\in \Sigma^{*}$,  if $(q_{1}, \ell_{1})\in \chi_{1}$, $(q_{2},
\ell_{2})\in \chi_{2}$, $\delta(q_{1}, s)=q_{2}$,
$\delta_{d}(\chi_{1}, P_{\sigma}(s))=\chi_{2}$, then
$F_{i}\in\ell_{1}$ implies $F_{i}\in\ell_{2}$.

\begin{proof} It can be directly verified from Definitions 6 and Definitions 7.
\end{proof}

{\it Property 2:}  If $\chi\in Q_{d}$,  then $(q_{1}, \ell_{1}),
\hskip 1mm (q_{2}, \ell_{2})\in \chi $ if and only if there exist
$s_{1}, s_{2}\in \Sigma^{*}$ such that $(s_{1})_{f}=(s_{2})_{f}\in
\Sigma_{d}$, $P_{\sigma}(s_{1})=P_{\sigma}(s_{2})$,
$\delta_{d}(\chi_{0}, P_{\sigma}(s_{1}))=\chi$, and for $k=1,2$,
${\cal L}(s_{k})>0$, $$\delta(q_{0}, s_{k})=q_{k}, \hskip 2mm
LP(q_{0}, \left\{N\right\}, s_{k} )= \ell_{k}.$$

\begin{proof}
{\it Necessity:}  If $\chi\in Q_{d}$, then there are
$a_{1},\ldots,a_{j}\in \Sigma_{d}$ and $\chi_{1},
\ldots,\chi_{j-1}\in Q_{d}$, such that $\delta_{d}(\chi_{i},
a_{i+1})=\chi_{i+1}$, where $0\leq i\leq j-1$ and $\chi_{j}=\chi$.
From the assumption that $(q_{1}, \ell_{1}), \hskip 1mm (q_{2},
\ell_{2}) \in \chi $, there exist $(q^{k}_{1}, \ell^{k}_{1})\in
\chi_{j-1}$,  and $t^{k}_{1}\in {\cal L}_{a_{j}}(q^{k}_{1}, \sigma
)$ ($k=1,2$) such that for $k=1,2$,
$$q_{k}=\delta(q^{k}_{1}, t^{k}_{1}), \hskip 4mm \ell_{k}=LP(q^{k}_{1}, \ell^{k}_{1}, t^{k}_{1}).$$ Similarly, note
that $\delta_{d}(\chi_{j-2}, a_{j-1})=\chi_{j-1}$, hence, there
are $(q^{k}_{2}, \ell^{k}_{2})\in \chi_{j-2}$, and $t^{k}_{2}\in
{\cal L}_{a_{j-1}}(q^{k}_{2}, \sigma )$ ($k=1,2$) satisfying  for
$k=1,2$,
$$q^{k}_{1}=\delta(q^{k}_{2}, t^{k}_{2}),\hskip 4mm \ell^{k}_{1}=LP(q^{k}_{2}, \ell^{k}_{2}, t^{k}_{2}).$$
$$\ldots\ldots $$ With the analogous process, there are $(q^{k}_{j-1}, \ell^{k}_{j-1})\in \chi_{1}$,
$t^{k}_{j}\in {\cal L}_{a_{1}}(q_{0}, \sigma )$ ($k=1,2$) such
that  for $k=1,2$,
$$q^{k}_{j-1}=\delta(q_{0}, t^{k}_{j}),\hskip 4mm  \ell^{k}_{j-1}=LP(q_{0},\left\{N\right\}, t^{k}_{j}).$$

We take \begin{equation} s_{k}=t^{k}_{j}t^{k}_{j-1}\ldots
t^{k}_{2} t^{k}_{1},\hskip 3mm (k=1,2).\end{equation} Obviously,
$\delta_{d}(\chi_{0}, P_{\sigma}(s_{1}))=\chi$,
$(s_{1})_{f}=(s_{2})_{f}=a_{j}\in \Sigma_{d}$ and for $k=1,2$, we
have ${\cal L}(s_{k})>0$, $\delta(q_{0}, s_{k})=q_{k}$, $LP(q_{0},
\left\{N\right\}, s_{k} )= \ell_{k}$. Moreover
$$P_{\sigma}(s_{1})=a_{1}a_{2}\ldots a_{j}=P_{\sigma}(s_{2}).$$

{\it Sufficiency:}  Assume that there exist $s_{1}, s_{2}\in
\Sigma^{*}$ satisfying ${\cal L}(s_{1})>0$, ${\cal L}(s_{2})>0$
and $ P_{\sigma}(s_{1})=P_{\sigma}(s_{2})$. From
$\delta_{d}(\chi_{0}, P_{\sigma}(s_{1}))=\chi$, we denote
$$P_{\sigma}(s_{1})=a_{1}a_{2}\ldots a_{j},$$  then we can obtain a
state sequence $\chi_{1}, \chi_{2},\ldots,\chi_{j-1}\in Q_{d}$
such that $\delta_{d}(\chi_{i}, a_{i+1})=\chi_{i+1}$, where $0\leq
i\leq j-1$ and $\chi_{j}=\chi$. Furthermore, from $\delta(q_{0},
s_{k})=q_{k}$, and $ LP(q_{0}, \left\{N\right\}, s_{k} )=
\ell_{k}$, ($k=1,2$), we have that $(q_{1}, \ell_{1}), \hskip 2mm
(q_{2}, \ell_{2})\in \chi $ by Definition 7. \end{proof}

{\it Remark 4:} In the proof of Necessity, it is possible that
$(q^{1}_{h}, \ell^{1}_{h})$ is the same as $(q^{2}_{h},
\ell^{2}_{h})$ for some $h$, but it does not concern the proof.

{\it Definition 8:}  Let $G_{d}=(Q_{d},\Sigma_{d},\delta_{d},
\chi_{0}) $ be the diagnoser with respect to $\sigma$. A state
$\chi\in Q_{d}$ is said to be {\it $F_{i}$-certain } if either
$F_{i}\in \ell$ for all $(q, \ell)\in \chi$, or $F_{i}\not\in\ell$
for all $(q, \ell)\in \chi$.  And $\chi$ is said to be {\it
$F_{i}$-uncertain }, if there are $(q_{1}, \ell_{1}), (q_{2},
\ell_{2})\in \chi$ such that $F_{i}\in \ell_{1}$  and
$F_{i}\not\in\ell_{2}$.

For example, $\chi_{1}=\{(q_{1},\{F_{2}\}),
(q_{5},\{F_{1},F_{2}\})\}$ and $\chi_{2}=\{(q_{2},\{F_{2}\}),
(q_{6},\{F_{1},F_{2}\})\}$ in Fig.8 are both $F_{2}$-certain and
$F_{1}$-uncertain states.

{\it Property 3:}  Let $G_{d}=(Q_{d},\Sigma_{d},\delta_{d},
\chi_{0}) $ be the diagnoser with respect to $\sigma$ and $
\delta_{d}(\chi_{0}, u)=\chi$.  If $\chi$ is $F_{i}$-certain, then
either $\widetilde{\Sigma}_{f_{i}}(s)\geq
\widetilde{\Sigma}_{f_{i}}(\sigma)$ for all $s\in
P_{\sigma}^{-1}(u)$, or
$\widetilde{\Sigma}_{f_{i}}(s)<\widetilde{\Sigma}_{f_{i}}(\sigma)$
for all $s\in P_{\sigma}^{-1}(u)$, where $s_{f}\in \Sigma_{d} $.

\begin{proof}  By contradiction, suppose there exist $s_{1}, s_{2}\in
P_{\sigma}^{-1}(u)$ such that
$$\widetilde{\Sigma}_{f_{i}}(s_{1})\geq \widetilde{\Sigma}_{f_{i}}(\sigma)
>\widetilde{\Sigma}_{f_{i}}(s_{2})$$
where $(s_{1})_{f}, (s_{2})_{f}\in\Sigma_{d} $. Denote $$LP(q_{0},
\left\{N\right\}, s_{1} )=\ell_{1}, \hskip 2mm LP(q_{0},
\left\{N\right\}, s_{2})=\ell_{2},$$ then from Definition 6, we
know that $F_{i}\in \ell_{1}$, but $F_{i}\not\in \ell_{2}$. By
Property 2, we have $(q_{1}, \ell_{1}), \hskip 2mm (q_{2},
\ell_{2})\in \chi $, where $ \delta(q_{0}, s_{1})=q_{1}$ and $
\delta(q_{0}, s_{2})=q_{2}$. That is, $\chi $ is
$F_{i}$-uncertain. \end{proof}

{\it Property 4:}  Let $G_{d}=(Q_{d},\Sigma_{d},\delta_{d},
\chi_{0}) $ be the diagnoser with respect to $\sigma$ and $
\delta_{d}(\chi_{0}, u)=\chi$.  If $\chi$ is $F_{i}$-uncertain,
then there exist $s_{1}, s_{2}\in \Sigma^{*}$ such that
$(s_{1})_{f}=(s_{2})_{f}\in\Sigma_{d} $, $ P_{\sigma}(s_{1})=
P_{\sigma}(s_{2})$, $\delta_{d}(\chi_{0},
P_{\sigma}(s_{1}))=\chi$, and
\begin{equation}\widetilde{\Sigma}_{f_{i}}(s_{1})\geq
\widetilde{\Sigma}_{f_{i}}(\sigma)>\widetilde{\Sigma}_{f_{i}}(s_{2}).\end{equation}

\begin{proof} It is straight obtained by Property 3. \end{proof}

{\it Property 5:}  Let $G_{d}=(Q_{d},\Sigma_{d},\delta_{d},
\chi_{0}) $ be the diagnoser with respect to $\sigma$. If the set
of states in $Q_{d}$ forms a cycle in $G_{d}$, then all states in
the cycle have the same failure label.

\begin{proof}  It is easy to prove since any two states in a cycle
of $G_{d}$ are reachable from each other, and once a failure label
is appended, it cannot be removed in all successors. \end{proof}

\subsection{Necessary and Sufficient Condition of Diagnosability for FDESs}
In this subsection, we present an approach of failure diagnosis in
the framework of FDESs, and a necessary and sufficient condition
of the diagnosability for FDESs is obtained.

We may define an $F_{i}$-indeterminate cycle in diagnosers for
FDESs, just as for classical DESs.

{\it Definition 9:}  Let $G_{d}=(Q_{d},\Sigma_{d},\delta_{d},
\chi_{0})$ be the diagnoser with respect to $\sigma$.  A set of
$F_{i}$-uncertain states $\chi_{1}, \chi_{2},\ldots,\chi_{k}\in
Q_{d}$ is said to form an {\it $F_{i}$-indeterminate cycle} if

(1) $\chi_{1}, \chi_{2}, \ldots,\chi_{k}$ form a cycle in $G_{d}$,
i.e., there is $\sigma_{j}\in\Sigma_{d}$ such that
$\delta_{d}(\chi_{j}, \sigma_{j})=\chi_{(j+1)\bmod k}$, where
$j=1,\ldots,k$.

(2) $\exists$ $(x_{j}^{h}, \ell_{j}^{h}), (y_{j}^{r},
d_{j}^{r})\in \chi_{j}$ ($j\in[1,k]$; $h\in[1,m]$; $r\in[1,n]$)
such that
\begin{enumerate}
\item $F_{i}\in \ell_{j}^{h}$ but $F_{i}\not\in
 d_{j}^{r}$ for all $ j, h, r;$
\item The sequences of states $\left\{x_{j}^{h}\right\}$ and
$\left\{y_{j}^{r}\right\}$ form cycles respectively in $G$  with
\begin{eqnarray*}
\delta(x_{j}^{h}, s_{j}^{h}\sigma_{j})=x_{j+1}^{h},
(j\in[1,k-1]; h\in[1,m]),\\
\delta(x_{k}^{h}, s_{k}^{h}\sigma_{k})=x_{1}^{h+1}, (h\in[1,m-1]),
\end{eqnarray*}
 and $\delta(x_{k}^{m}, s_{k}^{m}\sigma_{k})=x_{1}^{1}$;
 \begin{eqnarray}
 \nonumber \delta(y_{j}^{r}, t_{j}^{r}\sigma_{j})=y_{j+1}^{r},  (j\in[1,k-1];
r\in[1,n]),\\
\nonumber \delta(y_{k}^{r}, t_{k}^{r}\sigma_{k})=y_{1}^{r+1},
(r\in[1,n-1]),
\end{eqnarray}
and $\delta(y_{k}^{n}, t_{k}^{n}\sigma_{k})=y_{1}^{1}$,
\end{enumerate}
where $ s_{j}^{h}\sigma_{j}\in {\cal L}(x_{j}^{h}, \sigma)$, $
t_{j}^{r}\sigma_{j}\in {\cal L}(y_{j}^{r}, \sigma)$.

Intuitively,  an $F_{i}$-indeterminate cycle in $G_{d}$ is a cycle
composed of $F_{i}$-uncertain states where, corresponding to this
cycle, there exist two sequences $\left\{x_{j}^{h}\right\}$ and
$\left\{y_{j}^{r}\right\}$ forming cycles of $G$, in which  one
carries and the other does not carry failure label $F_{i}$.

Now we can present a necessary and sufficient condition of the
diagnosability for FDEs.

{\it Theorem 1:}  A fuzzy language ${\cal L}$ generated by a fuzzy
finite automaton $G$ is $F_{i}$-diagnosable if and only if for any
$\sigma\in \Sigma_{fail_{i}}$, the diagnoser $G_{d}$ with respect
to $\sigma$ satisfies the condition:  There are no
$F_{i}$-indeterminate cycles in $G_{d}$.

\begin{proof}
{\it Necessity:}  We prove it by contradiction. Assume that ${\cal
L}$ is $F_{i}$-diagnosable, and there is an $F_{i}$-indeterminate
cycle $\chi_{1}, \chi_{2},\ldots,\chi_{k}$ in diagnoser $G_{d}$
with respect to $\sigma$, where $\sigma\in \Sigma_{fail_{i}}$. By
Definition 9, the corresponding sequences of states
$\left\{x_{j}^{h}\right\}$ and $\left\{y_{j}^{r}\right\}$ form two
cycles in $G$, and the corresponding strings $s_{j}^{h}\sigma_{j}$
and $t_{j}^{r}\sigma_{j}$ satisfy condition 2) of Definition 9,
where $(x_{j}^{h}, \ell_{j}^{h}), (y_{j}^{r}, d_{j}^{r})\in
\chi_{j}$, and $F_{i}\in \ell_{j}^{h}$ but $F_{i}\not\in
 d_{j}^{r}$ for all $ j=1,\cdots,k$; $h=1,\cdots,m$; $r=1,\cdots,n$.

Since $(x_{1}^{1}, \ell_{1}^{1}), (y_{1}^{1}, d_{1}^{1})\in
\chi_{1}$, from Property 2, there exist $s_{0}, t_{0}\in
\Sigma^{*}$ such that $P_{\sigma}(s_{0})=P_{\sigma}(t_{0})$, $
\delta(q_{0}, s_{0})=x_{1}^{1}$, and $\delta(q_{0},
t_{0})=y_{1}^{1}$. Notice that $F_{i}\in \ell_{1}^{1}$ and
$F_{i}\not\in
 d_{j}^{r}$ for all $j,r$. Therefore, we have $\widetilde{\Sigma}_{f_{i}}(t_{0})<
\widetilde{\Sigma}_{f_{i}}(\sigma)$, and
\begin{equation}
 \widetilde{\Sigma}_{f_{i}}(s_{0})\geq
\widetilde{\Sigma}_{f_{i}}(\sigma)\geq\widetilde{\Sigma}_{f_{i}}(t_{j}^{r}\sigma_{j}).
\end{equation}

Let $l$ be arbitrarily large. We consider the following two traces
\begin{eqnarray}
\omega_{1}=s_{0}(s_{1}^{1}\sigma_{1}\ldots s_{k}^{1}\sigma_{k}
\ldots
s_{1}^{m}\sigma_{1}\ldots s_{k}^{m}\sigma_{k})^{ln},\\
\omega_{2}=t_{0} (t_{1}^{1}\sigma_{1}\ldots t_{k}^{1}\sigma_{k}
\ldots t_{1}^{n}\sigma_{1}\ldots t_{k}^{n}\sigma_{k})^{lm}.
\end{eqnarray}
Then ${\cal L}(\omega_{1})>0, \hskip 1mm{\cal L}(\omega_{2})>0$
and
\begin{eqnarray}
P_{\sigma}(\omega_{1})=P_{\sigma}(\omega_{2})=P_{\sigma}(s_{0})
(\sigma_{1}\sigma_{2}\ldots\sigma_{k})^{lmn}.\end{eqnarray}

Because $\widetilde{\Sigma}_{f_{i}}(s_{0})\geq
\widetilde{\Sigma}_{f_{i}}(\sigma)$, there is a prefix $s$ of
$s_{0}$ such that $s\in
\Psi_{\sigma}(\widetilde{\Sigma}_{f_{i}})$.  Take $t\in {\cal
L}/s$ where $\omega_{1}=st$, then from (25), we know $\omega_{2}
\in P_{\sigma}^{-1}(P_{\sigma}(st))$.  But from Ineqs.(22), and
\begin{eqnarray*}
\begin{array}{ll}
\widetilde{\Sigma}_{f_{i}}(\omega_{2})=\\
\max\{\widetilde{\Sigma}_{f_{i}}(t_{0}),
\widetilde{\Sigma}_{f_{i}}(t_{j}^{r}\sigma_{j}): j=1,\cdots,k;
r=1,\cdots,n\},
\end{array}\end{eqnarray*}
we have $\widetilde{\Sigma}_{f_{i}}(\omega_{2})<
\widetilde{\Sigma}_{f_{i}}(\sigma)$. That is, ${\cal L}$ is not
$F_{i}$-diagnosable, which contradicts the assumption.

{\it Sufficiency:}  Assume that there are no $F_{i}$-indeterminate
cycles in diagnoser $G_{d}$ with respect to $\sigma$, where
$\sigma\in \Sigma_{fail_{i}}$. The proof of sufficiency will be
completed by following two steps: (1) $\chi_{0}$ can reach an
$F_{i}$-certain state after a finite number of transitions; (2)
${\cal L}$ is $F_{i}$-diagnosable with respect to $\sigma$.

(1) Firstly, we verify that $\chi_{0}$ can reach an
$F_{i}$-certain state after a finite number of transitions.

For simplicity, if $(q, \ell), (q^{'}, \ell^{'})\in \chi $, and
$F_{i}\in \ell$, $F_{i}\not\in \ell^{'}$, we shall denote $q$ as
``$x$-state" of $\chi $ and $q^{'}$ as ``$y$-state" of $\chi $,
respectively. Let $s\in \Psi_{\sigma}(\widetilde{\Sigma}_{f_{i}})$
and $\delta(q_{0}, s)=q$. From Assumption (A2), there exists
$n_{0}\in N $ such that $\parallel t_{1}\parallel\leq n_{0}$ for
any $t_{1}\in {\cal L}(q, \sigma)$. Denote $\delta(q_{0},
st_{1})=q_{1}$, $\delta_{d}(\chi_{0},
P_{\sigma}(st_{1}))=\chi_{1}$, then $q_{1}$ is an ``$x$-state"
since $s\in \Psi_{\sigma}(\widetilde{\Sigma}_{f_{i}})$ implies
$\widetilde{\Sigma}_{f_{i}}(st_{1})\geq
\widetilde{\Sigma}_{f_{i}}(\sigma)$.

The desired result is obtained if $\chi_{1}$ is $F_{i}$-certain.
So the following is to prove the desired result under the
assumption that $\chi_{1}$ is $F_{i}$-uncertain. Since there are
no $F_{i}$-indeterminate cycles in $G_{d}$, one of the following
is true: (i) there are no cycles of $F_{i}$-uncertain states in
$G_{d}$, or (ii) there is one or more cycles of $F_{i}$-uncertain
states in $G_{d}$ but corresponding to such cycle, there do not
exist two sequences of ``$x$-states" and of ``$y$-states" forming
cycles in $G$.

{\it Case (i):} Suppose that there are no cycles of
$F_{i}$-uncertain states in $G_{d}$, which means $F_{i}$-uncertain
states will reach an $F_{i}$-certain state by Assumption (A1) and
Property 1. Therefore, there is sufficiently long $t_{2}\in {\cal
L}_{G}(q_{1})$ such that $\delta_{d}(\chi_{0},
P_{\sigma}(st_{1}t_{2} ))$ is an $F_{i}$-certain state.

{\it Case (ii):}  Suppose that there is a cycle of
$F_{i}$-uncertain states $\chi_{1}, \chi_{2},\ldots,\chi_{k}$ in
$G_{d}$, but correspondingly to such cycle, there do not exist two
sequences of ``$x$-states" and of ``$y$-states" forming cycles in
$G$. The following will prove that this case is impossible. In
fact, there is an ``$x$-state" $q_{2}$ of $\chi_{2}$ such that
$q_{2}$ is a successor of $q_{1}$ since $q_{1}$ is an ``$x$-state"
of $\chi_{1}$. Similarly, there is an ``$x$-state" $q_{3}$ of
$\chi_{3}$ such that $q_{3}$ is a successor of $q_{2}$. $\ldots$.
So, we obtain a sequence $\left\{q_{1},q_{2},\cdots\right\}$ of
``$x$-states" which forms cycles in $G$. With the analogous
process, we can obtain a sequence of ``$y$-states" which forms
cycles in $G$, too. That is, Case (ii) is impossible.

Above inference indicates that $\chi_{0}$ must reach an
$F_{i}$-certain state within a finite steps (denoted by $m_{0}$)
of transitions, no matter whether $\chi_{1}$ is $F_{i}$-certain or
not.

(2) From (1), we take $n_{i}=m_{0}$, then for any $s\in
\Psi_{\sigma}(\widetilde{\Sigma}_{f_{i}})$ and any $t\in {\cal
L}/s$ where $\parallel t \parallel\geq n_{i}$, $\chi_{0}$ must
lead to an $F_{i}$-certain state. That is, whenever $\omega \in
P_{\sigma}^{-1}(P_{\sigma}(st))$, it always holds that
$\widetilde{\Sigma}_{f_{i}}(\sigma)\leq
\widetilde{\Sigma}_{f_{i}}(\omega)$. Therefore, ${\cal L}$ is
$F_{i}$-diagnosable with respect to $\sigma$. \end{proof}

From the proof of Theorem 1, we know that Theorem 1 can be
precisely described as follows.

 {\it Theorem 2:}  A fuzzy language ${\cal L}$ generated by a
fuzzy finite automaton $G$ is $F_{i}-$diagnosable with respect to
$\sigma\in \Sigma_{fail_{i}}$ if and only if the diagnoser $G_{d}$
with respect to $\sigma$ satisfies the condition:  There are no
$F_{i}-$ indeterminate cycles in $G_{d}$.

\begin{proof}  It has been shown in the proof of Theorem 1. \end{proof}

\section{Examples of  Diagnosability for FDESs}\label{S:EXP}
In this section, we will give some examples to illustrate the
process of testing the necessary and sufficient condition for the
diagnosability of FDESs presented above, which may be viewed as an
applicable background of diagnosability for FDESs. Examples 2 and
3 are diagnosability for FDESs with single failure type: one is
diagnosable but the other is not diagnosable. Example 4 is
considered as an FDES with multiple failure types. For simplicity,
the fuzzy events (matrices) used are all upper or lower triangular
matrices.

{\it Example 2.} Let us use a fuzzy automaton $G=(Q, \Sigma,
\delta, q_{0})$ to model a patient's body condition. For
simplicity, we consider patient's condition roughly to be three
cases, i.e., ``poor" , ``fair", and ``excellent".  Suppose that
patient's initial condition (initial fuzzy state) is $q_{0}=[0.9,
\hskip 1mm 0.1 , \hskip 1mm 0]$, which means that the patient is
in a state with possibility of 0.9 for ``poor", 0.1 for ``fair"
and 0 for ``excellent". Suppose that there are three treatments to
choose for doctor, denoted as $\alpha$, $\beta$ and $\gamma$,
which are defined as follows:
$$
\alpha=\left[
\begin{array}{ccc}
0.4 &\ 0.9&\ 0.4\\
0&\ 0.4 &\ 0.4\\
0&\ 0&\ 0.4
\end{array}
\right], \beta=\left[
\begin{array}{ccc}
0.4 &\ 0&\ 0\\
0.9&\ 0.4&\ 0\\
0.4&\ 0.4&\ 0.4
\end{array}
\right],$$ $$ \gamma=\left[
\begin{array}{ccc}
0.9&\ 0.9&\ 0.4\\
0&\ 0.4&\ 0.4\\
0&\ 0&\ 0.4
\end{array}
\right].
$$

In general, it is possible that patient's condition turns better
or worse after each treatment, which may be evaluated by means of
experience and medical theory. For instance, fuzzy event $\alpha$
means that, after this treatment, the possibilities that patient's
status changes from ``poor" to ``poor", ``fair" and ``excellent"
are 0.4, 0.9 and 0.4; the possibilities from ``fair" to ``poor",
``fair" and ``excellent" are 0, 0.4 and 0.4; and the possibilities
from ``excellent" to ``poor", ``fair" and ``excellent" are 0, 0
and 0.4, respectively. Fuzzy events $\beta$ and $\gamma$ have
similar interpretations.

Assume that doctor's strategy for patient's treatment is described
by Fig.2. From $q_{0}=[0.9, \hskip 1mm 0.1 , \hskip 1mm 0]$, we
can calculate the other fuzzy states using the transition function
$\delta$ as: $q_{1}=[0.4, \hskip 1mm0.9,\hskip 1mm 0.4]$,
$$\begin{array}{ll} q_{2}=[0.9, \hskip 1mm0.4, \hskip 1mm0.4],&
 q_{3}=[0.9,\hskip 1mm 0.9,\hskip 1mm 0.4],\\ q_{4}=[0.4, \hskip 1mm0.1, \hskip 1mm0],
 & q_{5}=[0.4, \hskip 1mm0.4,\hskip 1mm 0.4].\end{array}$$

\setlength{\unitlength}{0.1cm}
\begin{picture}(60,40)

\put(10,18){\circle{5}\makebox(-10,0){$q_{0}$}}
\put(25,28){\circle{5}\makebox(-10,0){$q_{1}$}}
\put(40,28){\circle{5}\makebox(-10,0){$q_{2}$}}
\put(55,28){\circle{5}\makebox(-10,0){$q_{3}$}}
\put(25,8){\circle{5}\makebox(-10,0){$q_{4}$}}
\put(40,8){\circle{5}\makebox(-10,0){$q_{5}$}}

\put(13,18){\vector(1,1){9}} \put(13,18){\vector(1,-1){9}}
\put(28,28){\vector(1,0){9}} \put(43,28){\vector(1,0){9}}
\put(28,8){\vector(1,0){9}}

\qbezier(27,30)(43,40)(55,31) \put(29,31.8){\vector(-1,-1){2}}
\put(46.5,8){\circle{8}} \put(50.7,8){\vector(0,1){1}}

\put(16,25){\makebox(0,0)[c]{$\alpha$}}
\put(16,11){\makebox(0,0)[c]{$\beta$}}
\put(33,30){\makebox(0,0)[c]{$\beta$}}
\put(47,30){\makebox(0,0)[c]{$\gamma$}}
\put(45,37){\makebox(0,0)[c]{$\alpha$}}
\put(33,10){\makebox(0,0)[c]{$\alpha$}}
\put(53,8){\makebox(0,0)[c]{$\alpha$}}

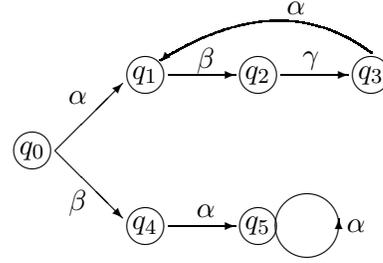
\put(25,1){\makebox(25,1)[c]{{\footnotesize Fig.2. The fuzzy
automaton of Example 2. }}}

\end{picture}

Fig.2 means that, if the patient obtains the first treatment being
$\alpha$ or $ \beta$, then his (or her) state changes into $q_{1}$
or $q_{4}$. After treatment $ \beta$ in condition $q_{1}$,
 the state will change from $q_{1}$ to $q_{2}$. And then, the
patient will turn into state $q_{3}$ after treatment $ \gamma$. If
treatment $ \alpha$ is adopted in state $q_{3}$, then the patient
returns to condition $q_{1}$. Similarly, when the patient obtains
treatment $ \alpha$ in $q_{4}$, the state will turn to $q_{5}$.
And the patient's condition will be unchanged if he or she obtains
treatment $ \alpha$ in $q_{5}$.

As mentioned above, for each treatment (fuzzy event), some effects
are observable, but some are unobservable, even if some are
undesired failures (for example, some potential side effects).
Therefore, each fuzzy event has certain degrees of observable and
unobservable, and, also, each fuzzy event may possess different
possibility of failure occurring. Assume that the degree of
observability and the possibility of failure occurring for each
fuzzy event are defined:
$$\begin{array}{lll}
\widetilde{\Sigma}_{o}(\alpha)=0.6, &
\widetilde{\Sigma}_{o}(\beta)=0.4, &
\widetilde{\Sigma}_{o}(\gamma)=0.7;\\
\widetilde{\Sigma}_{f_{1}}(\alpha)=0.1, &
\widetilde{\Sigma}_{f_{1}}(\beta)=0.2,&
\widetilde{\Sigma}_{f_{1}}(\gamma)=0.3.\end{array}
$$
 Now, in order to detect the occurrence of failure, we construct the diagnosers
with respect to each $\sigma\in \Sigma_{fail_{i}}$, where
$\Sigma_{fail_{i}}=\{\alpha, \beta, \gamma\}$.

(1). When $\sigma=\alpha$, the $\sigma$-projection $P_{\sigma}$ is
determined by $P_{\sigma}(\alpha)=P_{\sigma}(\beta)=\epsilon$,
$P_{\sigma}(\gamma)=\gamma$, and the set of events for the
diagnoser is $\Sigma_{d}=\{\gamma\}$. According to Definition 5,
the diagnoser $G_{d}$ with respect to $\alpha$ is constructed in
Fig.3. Obviously, there are no $F_{1}$-indeterminate cycles in
$G_{d}$. Therefore, by Theorem 2, ${\cal L}$ is
$F_{1}$-diagnosable with respect to $\alpha$. In fact, due to
$\widetilde{\Sigma}_{f_{1}}(\alpha)$ being the smallest among
$\{\widetilde{\Sigma}_{f_{1}}(a): a\in \Sigma\}$, Ineq.(13)
naturally holds with $n_{i}=0$.

\setlength{\unitlength}{0.1cm}
\begin{picture}(170,17)

\put(10,7){\framebox(10,6){$q_{0}N$}}
\put(35,7){\framebox(10,6){$q_{3}F_{1}$}}

\put(20,10){\vector(1,0){15}} \put(50,10){\circle{10}}
\put(55,10){\vector(0,1){1}}

\put(28,12){\makebox(0,0)[c]{$\gamma$}}
\put(58,10){\makebox(0,0)[c]{$\gamma$}}

\put(25,1){\makebox(25,1)[c]{{\footnotesize Fig.3. The diagnoser
$G_{d}$ w.r.t $\alpha$ in Example 2. }}}

\end{picture}

 (2). When $\sigma=\beta$, we have $P_{\sigma}(\alpha)=\alpha$,
$P_{\sigma}(\beta)=\epsilon$, $P_{\sigma}(\gamma)=\gamma$ and
$\Sigma_{d}=\{\alpha, \gamma\}$. And the diagnoser $G_{d}$ with
respect to $\beta$ is constructed in Fig.4. Obviously, ${\cal L}$
is $F_{1}$-diagnosable with respect to $\beta$ for no
$F_{1}$-indeterminate cycles in $G_{d}$. In fact, Ineq.(13) holds
with $n_{i}=1$.

(3). When $\sigma=\gamma$, we have
$P_{\sigma}(\alpha)=P_{\sigma}(\beta)=\epsilon$,
$P_{\sigma}(\gamma)=\gamma$ and $\Sigma_{d}=\{\gamma\}$. For no
$F_{1}$-indeterminate cycles in  the diagnoser $G_{d}$ with
respect to $\gamma$ constructed in Fig.5, ${\cal L}$ is
$F_{1}$-diagnosable with respect to $\gamma$.

Therefore, ${\cal L}$ is $F_{1}$-diagnosable. That is, the
occurrence of failure can be detected within finite delay.

\setlength{\unitlength}{0.1cm}
\begin{picture}(170,33)

\put(5,22){\framebox(10,6){$q_{0}N$}}
\put(25,22){\framebox(16,6){$q_{1}Nq_{5}F_{1}$}}
\put(50,22){\framebox(10,6){$q_{5}F_{1}$}}
\put(25,7){\framebox(10,6){$q_{3}F_{1}$}}
\put(45,7){\framebox(10,6){$q_{1}F_{1}$}}

\put(15,25){\vector(1,0){10}} \put(41,25){\vector(1,0){9}}
\put(30,22){\vector(0,-1){9}} \put(35.5,11){\vector(1,0){9.5}}
\put(44.5,9){\vector(-1,0){9.5}}

\put(64.3,25){\circle{8}} \put(68.5,25){\vector(0,1){1}}

\put(20,27){\makebox(0,0)[c]{$\alpha$}}
\put(45,27){\makebox(0,0)[c]{$\alpha$}}
\put(71,24){\makebox(0,0)[c]{$\alpha$}}
\put(27,17){\makebox(0,0)[c]{$\gamma$}}
\put(40,13){\makebox(0,0)[c]{$\alpha$}}
\put(40,7){\makebox(0,0)[c]{$\gamma$}}

\put(25,1){\makebox(20,1)[c]{{\footnotesize Fig.4. The diagnoser
$G_{d}$ w.r.t $\beta$ in Example 2. }}}

\end{picture}

\begin{picture}(170,20)

\put(10,7){\framebox(10,6){$q_{0}N$}}
\put(35,7){\framebox(10,6){$q_{3}F_{1}$}}

\put(20,10){\vector(1,0){15}} \put(50,10){\circle{10}}
\put(55,10){\vector(0,1){1}}

\put(28,12){\makebox(0,0)[c]{$\gamma$}}
\put(58,10){\makebox(0,0)[c]{$\gamma$}}

\put(25,1){\makebox(25,1)[c]{{\footnotesize Fig.5. The diagnoser
$G_{d}$ w.r.t $\gamma$ in Example 2. }}}

\end{picture}

{\it Example 3.} Consider the fuzzy automaton $G=(Q, \Sigma,
\delta, q_{0})$ represented in Fig.6, where  $Q=\{q_{0},
q_{1},\ldots, q_{7}\}$ is defined as: $$\begin{array}{ll}
q_{0}=[0.9,\hskip 1mm0.1,\hskip 1mm0],& q_{1}=[0.4, \hskip 1mm0.9,\hskip 1mm 0.4],\\
q_{2}=[0.9, \hskip 1mm0.4,\hskip 1mm0.4],& q_{3}=[0.9, \hskip 1mm0.9,\hskip 1mm0.4],\\
q_{4}=[0.5,\hskip 1mm0.1,\hskip 1mm0],&
 q_{5}=[0.4,\hskip 1mm 0.5,\hskip 1mm 0.4],\\ q_{6}=[0.5,\hskip 1mm 0.4,\hskip 1mm 0.4],&
 q_{7}=[0.5, \hskip 1mm0.5,\hskip 1mm 0.4].\end{array}$$

\setlength{\unitlength}{0.1cm}
\begin{picture}(60,38)

\put(5,18){\circle{5}\makebox(-10,0){$q_{0}$}}
\put(20,28){\circle{5}\makebox(-10,0){$q_{1}$}}
\put(35,28){\circle{5}\makebox(-10,0){$q_{2}$}}
\put(50,28){\circle{5}\makebox(-10,0){$q_{3}$}}
\put(20,8){\circle{5}\makebox(-10,0){$q_{4}$}}
\put(35,8){\circle{5}\makebox(-10,0){$q_{5}$}}
\put(50,8){\circle{5}\makebox(-10,0){$q_{6}$}}
\put(62,8){\circle{5}\makebox(-10,0){$q_{7}$}}

\put(8,18){\vector(1,1){9}} \put(8,18){\vector(1,-1){9}}
\put(23,28){\vector(1,0){9}} \put(38,28){\vector(1,0){9}}
\put(23,8){\vector(1,0){9}} \put(38,8){\vector(1,0){9}}
\put(53,8){\vector(1,0){6}}

\qbezier(22,30)(38,40)(49,31) \put(24,31.8){\vector(-1,-1){2}}
\qbezier(37,10)(48,20)(61,11) \put(39,11.8){\vector(-1,-1){2}}

\put(11,25){\makebox(0,0)[c]{$\alpha$}}
\put(11,11){\makebox(0,0)[c]{$\tau$}}
\put(28,30){\makebox(0,0)[c]{$\beta$}}
\put(42,30){\makebox(0,0)[c]{$\gamma$}}
\put(35,37){\makebox(0,0)[c]{$\alpha$}}
\put(28,10){\makebox(0,0)[c]{$\alpha$}}
\put(42,10){\makebox(0,0)[c]{$\beta$}}
\put(55,10){\makebox(0,0)[c]{$\gamma$}}
\put(50,18){\makebox(0,0)[c]{$\alpha$}}

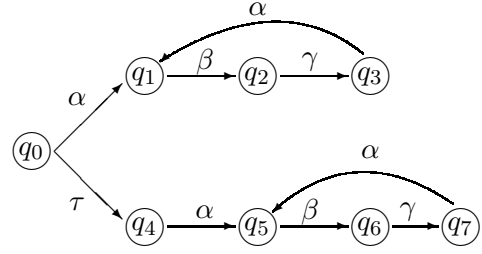
\put(25,1){\makebox(25,1)[c]{{\footnotesize Fig.6. The fuzzy
automaton of Example 3. }}}

\end{picture}

The set of fuzzy events $\Sigma=\{\tau,\alpha, \beta, \gamma\}$,
where $\tau,\alpha, \beta, \gamma$ are defined as follows:
$$\tau=\left[
\begin{array}{ccc}
0.5 & 0& 0\\
0.1& 0.1 & 0\\
0.1& 0.1& 0.1
\end{array}
\right],
 \alpha=\left[
\begin{array}{ccc}
0.4 & 0.9& 0.4\\
0& 0.4 & 0.4\\
0& 0& 0.4
\end{array}
\right],$$$$ \beta=\left[
\begin{array}{ccc}
0.4 & 0& 0\\
0.9& 0.4& 0\\
0.4& 0.4& 0.4
\end{array}
\right], \gamma=\left[
\begin{array}{ccc}
0.9& 0.9& 0.4\\
0& 0.4& 0.4\\
0& 0& 0.4
\end{array}
\right].
$$

Suppose that $\widetilde{\Sigma}_{o}$ and
$\widetilde{\Sigma}_{f_{1}}$ are defined as follows:
$$\begin{array}{ccc} \widetilde{\Sigma}_{o}(\tau)=0.3,
&\widetilde{\Sigma}_{o}(\alpha)=0.5, &
\widetilde{\Sigma}_{o}(\beta)=0.4,\\
\widetilde{\Sigma}_{o}(\gamma)=0.6;
&\widetilde{\Sigma}_{f_{1}}(\tau)=0.4,
&\widetilde{\Sigma}_{f_{1}}(\alpha)=0.1,\\
\widetilde{\Sigma}_{f_{1}}(\beta)=0.2,
&\widetilde{\Sigma}_{f_{1}}(\gamma)=0.3.\end{array}$$

We can verify that the language ${\cal L}$ is not
$F_{1}$-diagnosable. In fact, when $\sigma=\tau$, for arbitrary
$n_{i}\in N $, we take $s=\tau$, \hskip 1mm
$t=\alpha(\beta\gamma\alpha)^{n_{i}}$, and
$\omega=\alpha(\beta\gamma\alpha)^{n_{i}}$, and then $\omega \in
P_{\sigma}^{-1}(P_{\sigma}(st))$, but
$$\widetilde{\Sigma}_{f_{1}}(\sigma)=0.4 >
0.3\geq\widetilde{\Sigma}_{f_{1}}(\omega).$$ Therefore, by
Definition 4, we know that ${\cal L}$ is not $F_{1}$-diagnosable
with respect to $\tau$. Of course, the result can also be obtained
by the diagnoser $G_{d}$ with respect to $\tau$, which is
constructed in Fig.7, since there does exist an
$F_{1}$-indeterminate cycle in $G_{d}$.

\begin{picture}(60,30)

\put(5,7){\framebox(8,6){$q_{0}N$}}
\put(19,7){\framebox(16,6){$q_{1}Nq_{5}F_{1}$}}
\put(42,7){\framebox(16,6){$q_{2}Nq_{6}F_{1}$}}
\put(42,20){\framebox(16,6){$q_{3}Nq_{7}F_{1}$}}

\put(13,10){\vector(1,0){6}} \put(35,10){\vector(1,0){6}}
\put(50,13){\vector(0,1){7}} \put(42,22){\vector(-2,-1){17}}

\put(16,12){\makebox(0,0)[c]{$\alpha$}}
\put(38,12){\makebox(0,0)[c]{$\beta$}}
\put(52,16){\makebox(0,0)[c]{$\gamma$}}
\put(30,19){\makebox(0,0)[c]{$\alpha$}}

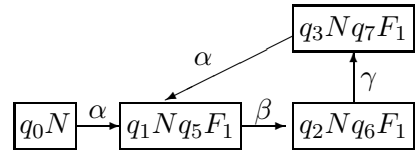
\put(22,1){\makebox(22,1)[c]{{\footnotesize Fig.7. The diagnoser
$G_{d}$ w.r.t $\tau$ in Example 3. }}}

\end{picture}

The following is an example of diagnosability for an FDES with
multiple failure types.

{\it Example 4.} Consider the fuzzy automaton $G=(Q, \Sigma,
\delta, q_{0})$ described in Example 3. The definition of
$\widetilde{\Sigma}_{o}$ is the same as that in Example 3, but
$\widetilde{\Sigma}_{f}=\widetilde{\Sigma}_{f_{1}}
\widetilde{\cup}\widetilde{\Sigma}_{f_{2}}$, which is defined as
follows:
$$\begin{array}{ccc}\widetilde{\Sigma}_{f_{1}}(\tau)=0.4,
&\widetilde{\Sigma}_{f_{1}}(\alpha)=0.1, &
\widetilde{\Sigma}_{f_{1}}(\beta)=0.2,\\
\widetilde{\Sigma}_{f_{1}}(\gamma)=0.3;&\widetilde{\Sigma}_{f_{2}}(\tau)=0.1,
&\widetilde{\Sigma}_{f_{2}}(\alpha)=0.2, \\
\widetilde{\Sigma}_{f_{2}}(\beta)=0.3,&
\widetilde{\Sigma}_{f_{2}}(\gamma)=0.4.\end{array}$$

The following is to verify that ${\cal L}$ is not $
F_{1}$-diagnosable but $ F_{2}$-diagnosable through constructing
the diagnosers.

(1). If $\sigma=\tau$, then $P_{\sigma}(\tau)=\epsilon$,
$P_{\sigma}(\alpha)=\alpha$, $P_{\sigma}(\beta)=\beta$,
$P_{\sigma}(\gamma)=\gamma$ and $\Sigma_{d}=\{\alpha, \beta,
\gamma\}$. Note that in the diagnoser $G_{d}$ with respect to
$\tau$ constructed as Fig.8, there exists an $F_{1}$-indeterminate
cycle but there do not exist $F_{2}$-indeterminate cycles.
Therefore, ${\cal L}$ is not $ F_{1}$-diagnosable but $
F_{2}$-diagnosable with respect to $\tau$. Of course, this result
can be verified by Definition 4, too. For failure type $f_{1}$, we
take $s=\tau$, \hskip 1mm $t=\alpha(\beta\gamma\alpha)^{n_{i}}$
and $\omega=\alpha(\beta\gamma\alpha)^{n_{i}}$, then $\omega \in
P_{\sigma}^{-1}(P_{\sigma}(st))$, but
$$\widetilde{\Sigma}_{f_{1}}(\sigma)=0.4
> 0.3\geq\widetilde{\Sigma}_{f_{1}}(\omega).$$ For
failure type $f_{2}$, since $\widetilde{\Sigma}_{f_{2}}(\tau)$ is
the least among $\{\widetilde{\Sigma}_{f_{2}}(a): a\in \Sigma\}$,
Ineq.(13) holds with $n_{i}=0$.

\setlength{\unitlength}{0.1cm}
\begin{picture}(65,28)

\put(6,7){\framebox(8,6){$q_{0}N$}}
\put(20,7){\framebox(20,6){$q_{1}F_{2}q_{5}F_{1}F_{2}$}}
\put(46,7){\framebox(21,6){$q_{2}F_{2}q_{6}F_{1}F_{2}$}}
\put(46,18){\framebox(21,6){$q_{3}F_{2}q_{7}F_{1}F_{2}$}}

\put(14,10){\vector(1,0){6}} \put(40,10){\vector(1,0){6}}
\put(57,13){\vector(0,1){5}} \put(46,21){\vector(-2,-1){15}}

\put(17,12){\makebox(0,0)[c]{$\alpha$}}
\put(43,12){\makebox(0,0)[c]{$\beta$}}
\put(59,15){\makebox(0,0)[c]{$\gamma$}}
\put(36,18){\makebox(0,0)[c]{$\alpha$}}

\put(25,1){\makebox(25,1)[c]{{\footnotesize Fig.8. The diagnoser
$G_{d}$ w.r.t $\tau$ in Example 4. }}}

\end{picture}

(2). If $\sigma=\alpha$, then
$P_{\sigma}(\tau)=P_{\sigma}(\alpha)=P_{\sigma}(\beta)=\epsilon$,
$P_{\sigma}(\gamma)=\gamma$ and $\Sigma_{d}=\{\gamma\}$. Note that
there do not exist $F_{1}$-indeterminate cycles or $F_{2}-$
indeterminate cycles in the diagnoser  with respect to $\alpha$
constructed in Fig.9, and ${\cal L}$ is both $ F_{1}$-diagnosable
and $F_{2}$-diagnosable with respect to $\alpha$. In fact,
Ineq.(13) holds for failure type $f_{1}$ with $n_{i}=0$ and for
$f_{2}$ with $n_{i}=2$.

\setlength{\unitlength}{0.1cm}
\begin{picture}(75,16)

\put(13,7){\framebox(10,6){$q_{0}N$}}
\put(30,7){\framebox(25,6){$q_{3}F_{1}F_{2}q_{7}F_{1}F_{2}$}}

\put(23,10){\vector(1,0){6}} \put(60,10){\circle{10}}
\put(65,10){\vector(0,1){1}}

\put(28,12){\makebox(0,0)[c]{$\gamma$}}
\put(68,10){\makebox(0,0)[c]{$\gamma$}}

\put(27,1){\makebox(27,1)[c]{{\footnotesize Fig.9. The diagnoser
$G_{d}$ w.r.t $\alpha$ in Example 4. }}}

\end{picture}

(3). If $\sigma=\beta$, then
$P_{\sigma}(\tau)=P_{\sigma}(\beta)=\epsilon$,
$P_{\sigma}(\alpha)=\alpha$, $P_{\sigma}(\gamma)=\gamma$, and
$\Sigma_{d}=\{\alpha, \gamma\}$.  There do not exist
$F_{1}$-indeterminate cycles or $F_{2}$-indeterminate cycles in
the diagnoser with respect to $\beta$, which is constructed as
Fig.10,   so ${\cal L}$ is both $F_{1}$-diagnosable and
$F_{2}$-diagnosable with respect to $\beta$. In fact, Ineq.(13)
holds for failure types $f_{1}$ and $f_{2}$ with $n_{i}=1$.

\setlength{\unitlength}{0.1cm}
\begin{picture}(170,30)

\put(6,7){\framebox(8,6){$q_{0}N$}}
\put(21,7){\framebox(15,6){$q_{1}Nq_{5}F_{1}$}}
\put(42,7){\framebox(25,6){$q_{3}F_{1}F_{2}q_{7}F_{1}F_{2}$}}
\put(42,20){\framebox(25,6){$q_{1}F_{1}F_{2}q_{5}F_{1}F_{2}$}}

\put(14,10){\vector(1,0){6}} \put(36,10){\vector(1,0){5}}
\put(55,13){\vector(0,1){7}}\put(42,22){\vector(-2,-1){17}}

\put(17,12){\makebox(0,0)[c]{$\alpha$}}
\put(38.5,12){\makebox(0,0)[c]{$\gamma$}}
\put(58,16){\makebox(0,0)[c]{$\alpha$}}
\put(32,19){\makebox(0,0)[c]{$\gamma$}}

\put(26,1){\makebox(26,1)[c]{{\footnotesize Fig.10. The diagnoser
$G_{d}$ w.r.t $\beta$ in Example 4. }}}
\end{picture}

(4). If $\sigma=\gamma$, then
$P_{\sigma}(\tau)=P_{\sigma}(\alpha)=P_{\sigma}(\beta)=\epsilon$,
$P_{\sigma}(\gamma)=\gamma$, and $\Sigma_{d}=\{\gamma\}$. Since
there do not exist $F_{1}$-indeterminate cycles or $F_{2}-$
indeterminate cycles in the diagnoser with respect to $\gamma$
constructed in Fig.11, ${\cal L}$ is both $F_{1}$-diagnosable and
$F_{2}$-diagnosable with respect to $\gamma$. In fact, Ineq.(13)
holds for failure type $f_{1}$ with $n_{i}=3$ and for $f_{2}$ with
$n_{i}=0$.

Therefore, by Theorem 1, we know that ${\cal L}$ is not
$F_{1}$-diagnosable but $F_{2}$-diagnosable.

\begin{picture}(170,19)
\put(15,7){\framebox(10,6){$q_{0}N$}}
\put(32,7){\framebox(25,6){$q_{3}F_{1}F_{2}q_{7}F_{1}F_{2}$}}

\put(25,10){\vector(1,0){6}} \put(62,10){\circle{10}}
\put(67,10){\vector(0,1){1}}

\put(28,12){\makebox(0,0)[c]{$\gamma$}}
\put(70,10){\makebox(0,0)[c]{$\gamma$}}

\put(26,1){\makebox(26,1)[c]{{\footnotesize Fig.11. The diagnoser
$G_{d}$ w.r.t $\gamma$ in Example 4. }}}

\end{picture}

\section{Concluding Remarks}
In this paper, we dealt with the diagnosability in the framework
of FDESs. We formalized the definition of diagnosability for
FDESs, in which the observable set and the failure set of events
are fuzzy. Then we constructed the observability-based diagnosers
and investigated its some basic properties. In particular, we
presented a necessary and sufficient condition for diagnosability
of FDESs. Our results generalized the important consequences in
classical DESs introduced by Sampath {\it et al} [30,31].
Moreover, the approach proposed in this paper may better deal with
the problems of fuzziness, impreciseness and subjectivity in the
failure diagnosis. As well, some examples serving to illuminate
the applications of the diagnosability of FDESs were described.

As pointed out above, FDESs have been applied to biomedical
control for HIV/AIDS treatment planning by Lin {\it et al} [20,21]
and also to intelligent sensory information processing for
robotics by R. Huq {\it et al} recently [10, 11]. The potential of
applications of the results in this paper may be used in those
systems. Moreover, with the results obtained in this paper, a
further issue worthy of consideration is the $I$-diagnosability
and the $AA$-diagnosability of FDESs,  as those investigated in
the frameworks of DESs [30] and stochastic DESs [36]. Another
important issue is how to detect the failures in decentralized
FDESs. Furthermore, FDESs modeled by fuzzy Petri nets [22] still
have not been dealt with. We would like to consider them in
subsequent work.


\begin{thebibliography}{AB}

\bibitem{ab}S.Bavshi and E. Chong,
``Automated fault diagnosis of using a discrete event systems
framework," in {\it Proc. 9th IEEE int. Symp. Intelligent Contr.},
1994, pp. 213-218.

\bibitem{ab}Y. Cao and M. Ying,  ``Supervisory control of fuzzy discrete event
systems,"
 {\it IEEE Trans.  Syst., Man,
Cybern. B}, vol. 35, no. 2, pp. 366-371, Apr. 2005.

\bibitem{ab}Y. Cao and M. Ying,
``Observability and Decentralized Control of Fuzzy Discrete-Event
Systems," {\it IEEE Trans.  Fuzzy Syst.}, vol. 14, no. 2, pp.
202-216, 2006.



\bibitem{ab} C. G. Cassandras and S. Lafortune, {\it Introduction to Discrete Event Systems.} Boston, MA: Kluwer, 1999.

\bibitem{ab}A. Darwiche and G. Provan,
``Exploiting system structure in modelbased diagnosis of discrete
event systems,"
 in {\it Proc. 7th Annu. Int. Workshop on the Principles of
 Diagnosis (DX'96)}, Oct. 1996, pp. 95-105.


\bibitem{ab}R. Debouk, ``Failure diagnosis of decentralized discrete event
systems,"  Ph.D. dissertation, Elec. Eng. Comp. Sci. Dept.,
University of Michigan, Ann Arbor, MI, 2000.

\bibitem{ab}R.Debouk, S. Lafortune, and D. Teneketzis,``On the effect of communication delays
in failure diagnosis of decentralized discrete event systems,"
 {\it Discrete Event Dyna. Syst.: Theory  Appl.}, 13(2003), pp.
 263-289.

\bibitem{ab}P. Frank, ``Fault diagnosis in dynamic systems using analytical
and knowledge based redundancy-A survey and some new results,"
 {\it  Automatica}, vol. 26, pp. 459-474,
 1990.

\bibitem{ab}E. Garcia, F. Morant, R. Blasco-Giminez, A. Correcher, and E. Quiles,
``Centralized modular diagnosis and the phenomenon of coupling,"
 in {\it Proc. 2002 IEEE Int. Workshop on Discrete
 Event Systems (WODES'02)}, Oct. 2002, pp. 161-168.

\bibitem{ab}R. Huq, G. K. I. Mann and R. G. Gosine,
``Distributed fuzzy discrete event system for robotic sensory
information processing," {\it Expert Systems}, vol. 23, no. 5, pp.
273-289, Nov. 2006.

\bibitem{ab}R. Huq, G. K. I. Mann and R. G. Gosine,
``Behavior-Madulation Technique in Mobile Robotics Using Fuzzy
Discrete Event System," {\it IEEE Trans. Robotics}, vol. 22, no.
5, pp. 903-916, Oct. 2006.



\bibitem{ab}S. Jiang and R. Kumar,
``Failure diagnosis of discrete event systems with linear-time
temporal logic fault specifications,"
 in {\it Proc. 2002 Amer. Control Conf.}, May 2002, pp. 128-133.

\bibitem{ab}S. Jiang, R. Kumar, and H. Garcia,
``Diagnosis of repeated failures in discrete event systems,"
 in {\it Proc. 41st IEEE Conf. Decision and Control}, Dec. 2002, pp. 4000-4005.

\bibitem{ab}G. J. Klir and B. Yuan,
{\it Fuzzy Sets and Fuzzy Logic: Theory and Applications.}
Englewood Cliffs, NJ: Prentice-Hall, 1995.

\bibitem{ab}S. Lafortune, D. Teneketzis, M. Sampath, R. Sengupta, and K. Sinnamohideen,
``Failure diagnosis of dynamic systems: An approach based on
discrete-event systems,"
 in {\it Proc.2001 Amer. Control Conf.}, Jun. 2001, pp. 2058-2071.

\bibitem{ab}G. Lamperti and M. Zanella,
``Diagnosis of discrete event systems integrating synchronous and
asynchronous behavior,"
 in {\it Proc. 9th Int. Workshop on Principles of Diagnosis (DX'99)}, 1999, pp. 129-139.

\bibitem{ab}F. Lin,
``Diagnosability of discrete event systems and its applications,"
{\it Discrete Event Dyna. Syst.: Theory Appl.}, vol. 4, no. 2, pp.
197-212,
 May. 1994.

\bibitem{ab}F. Lin and H. Ying, ``Fuzzy discrete event systems and their observability," in
{\it Pro. Joint Int. Conf. 9th Int. Fuzzy Systems Assoc. World
Congr. 20th North Amer. Fuzzy Inform. Process. Soci.}, Vancouver,
BC, Canada, July 25-28, 2001.

\bibitem{ab}F. Lin and H. Ying, ``Modeling and control of fuzzy discrete event systems," {\it IEEE Trans.  Syst., Man,
Cybern. B}, vol. 32, no. 4, Aug. pp. 408-415, 2002.

\bibitem{ab}F. Lin, H. Ying, X. Luan, R.D. MacArthur, J.A. Cohn, D.C. Barth-Jones, and L.R. Crane,
``Fuzzy discrete event systems and its applications to clinical
treatment planning," in {\it Proceedings of the 43rd IEEE Conf.
Decision and Control}, Budapest, Hungary, June 25-29, 2004, pp.
197-202.

\bibitem{ab}F. Lin, H. Ying, X. Luan, R.D. MacArthur, J.A. Cohn, D.C. Barth-Jones, and L.R. Crane,
``Theory for a control architecture of  fuzzy discrete event
system for decision making," in {\it 44th Conference on Decision
and Control and European Control Conference ECC}, 2005.



\bibitem{ab}C. G. Looney, ``Fuzzy petri nets for rule-based decisionmaking,"
{\it IEEE Trans.  Syst., Man, Cybern. B}, vol. 18, no. 1, pp.
178-183, 1988.

\bibitem{ab}J. Lunze and J. Schr\"{o}der, ``State observation and Diagnosis of
discrete-event systems described by stochastic automata,"
 {\it Discrete Event Dyna. Syst.: Theory Appl.}, vol. 11, pp.
 319-369, 2001.


\bibitem{ab}D. Pandalai and L. Holloway,
``Template languages for fault monitoring of discrete event
processes," {\it IEEE Trans. Automat. Contr.}, vol. 45, no. 5, pp.
868-882,
 May 2000.

\bibitem{ab}Y. Pencol\'{e},
``Decentralized diagnoser approach: Application to
telecommunication networks,"
 in {\it Proc. 11th Int. Workshop on Principles of
 Diagnosis (DX'00)}, Jun. 2000, pp. 185-192.

\bibitem{ab}G. Provan and Y.-L. Chen,
``Diagnosis of timed discrete event systems using temporal causal
networks: Modeling and analysis,"
 in {\it Proc. 1998 Int. Workshop on Discrete Event Systems (WODES'98)}, Aug. 1998, pp. 152-154.

\bibitem{ab}G. Provan and Y.-L. Chen,
``Model-based diagnosis and control reconfiguration for discrete
event systems: An integrated approach,"
 in {\it Proc. 38th IEEE Conf. Decision and Control}, Dec. 1999, pp. 1762-1768.


\bibitem{ab}D. W. Qiu,  ``Supervisory control of fuzzy discrete event systems: A formal approach,"
 {\it IEEE Trans.  Syst., Man,
Cybern. B}, vol. 35, no. 1, pp. 72-88, Feb. 2005.

\bibitem{ab}L. Roz\'{e} and M. O. Cordier,``Diagnosing discrete event systems:
Extending the ``Diagnoser Approach" to deal with telecommunication
networks," {\it Discrete Event Dyna. Syst.: Theory Appl.}, vol.
12, pp.
 43-81, 2002.

\bibitem{ab}M. Sampath, S. Lafortune, and D. Teneketzis,
``Active diagnosis of discrete-event systems," {\it IEEE Trans.
Automat. Contr.}, vol. 43, no. 7, pp. 908-929,
 Jul. 1998.

\bibitem{ab}M. Sampath, R. Sengupta, S. Lafortune, K. Sinnamohideen, and D. Teneketzis,
``Diagnosability of discrete-event systems,"
 {\it IEEE Trans. Automat. Contr.}, vol. 40, no. 9, pp. 1555-1575,
 Sep. 1995.

\bibitem{ab}M. Sampath, R. Sengupta, S. Lafortune, K. Sinnamohideen, and D. Teneketzis, ``Failure diagnosis
using discrete-event models,"
 {\it IEEE Trans. Automat. Contr. Syst. Technol.}, vol. 4, no. 2, pp. 105-124,
 Mar. 1996.

\bibitem{ab}R. Sengupta,
``Discrete-event diagnostics of automated vehicles and highways,"
 in {\it Proc. 2001 Amer. Control Conf.}, Jun. 2001.

\bibitem{ab}K. Sinnamohideen,
``Discrete-event diagnostics of heating, ventilation, and
air-conditioning systems,"
 in {\it Proc. 2001 Amer. Control Conf.}, Jun. 2001, pp.
 2072-2076.

\bibitem{ab}R. Su and W. Wonham,
``Global and local consistencies in distributed fault diagnosis
for discrete-event systems,"
 {\it IEEE Trans. Automat. Contr.}, vol. 50, no. 12, pp. 1923-1935,
 Dec. 2005.

\bibitem{ab}D. Thorsley, and D. Teneketzis, ``Diagnosability of stochastic discrete-event systems,"
 {\it IEEE Trans. Automat. Contr.}, vol. 50, no. 4, pp. 476-492,
 April. 2005.

\bibitem{ab}N. Viswanadham and T. Johnson,
``Fault detection and diagnosis of automated manufacturing
systems,"
 in {\it Proc. 27th IEEE Conf. Decision and Control}, Dec. 1988, pp. 2301-2306.

\bibitem{ab}G. Westerman, R. Kumar, C. Stround, and J. Heath,
``Discrete event system approach for delay fault analysis in
digital circuits,"
 in {\it Proc. 1998 Amer. Control Conf.}, Jun. 1998, pp.
 239-243.

\bibitem{ab}S. H. Zad, R. Kwong, and W. Wonham,
``Fault diagnosis in discrete event systems: Framework and model
reduction,"
 in {\it Proc. 37th IEEE Conf. Decision and Control}, Dec. 1998, pp. 3769-3774.



\bibitem{ab}L. A. Zadeh,  ``Fuzzy Logic=Computing with Words,"
{\it IEEE Trans.  Fuzzy Syst.}, vol. 4, no. 2, pp. 103-111, 1996.




\end{thebibliography}
\end{document}